\documentclass[journal]{IEEEtran}
\ifCLASSINFOpdf
  \usepackage[pdftex]{graphicx}
  \graphicspath{{./}}
  \usepackage{epstopdf}
\else
\fi
\usepackage{amsmath}
\usepackage{array}
\ifCLASSOPTIONcompsoc
  \usepackage[caption=false,font=normalsize,labelfont=sf,textfont=sf]{subfig}
\else
  \usepackage[caption=false,font=footnotesize]{subfig}
\fi
\begin{document}
%
% paper title
% Titles are generally capitalized except for words such as a, an, and, as,
% at, but, by, for, in, nor, of, on, or, the, to and up, which are usually
% not capitalized unless they are the first or last word of the title.
% Linebreaks \\ can be used within to get better formatting as desired.
% Do not put math or special symbols in the title.
%\title{A Keypoint-based and Unified Framework for 3D Hand Pose Estimation Using Monocular RGB Images}
%\title{Unconstrained Monocular 3D Hand Pose Estimation by keypoint detection with geometry constraints}
\title{An End-to-end Framework for Unconstrained Monocular 3D Hand Pose Estimation}
%
% author names and IEEE memberships
% note positions of commas and nonbreaking spaces ( ~ ) LaTeX will not break
% a structure at a ~ so this keeps an author's name from being broken across
% two lines.
% use \thanks{} to gain access to the first footnote area
% a separate \thanks must be used for each paragraph as LaTeX2e's \thanks
% was not built to handle multiple paragraphs
%

\author{Sanjeev~Sharma,
        Shaoli~Huang,and~Dacheng~Tao \\
        University of Sydney}% <-this % stops a space

\maketitle

% As a general rule, do not put math, special symbols or citations
% in the abstract or keywords.
\begin{abstract}
This work addresses the challenging problem of unconstrained 3D hand pose estimation using monocular RGB images. Most of the existing approaches assume some prior knowledge of hand (such as hand locations and side information) is available for 3D hand pose estimation. This restricts their use in unconstrained environments. We, therefore, present an end-to-end framework that robustly predicts hand prior information and accurately infers 3D hand pose by learning \textit{ConvNet} models while only using keypoint annotations. To achieve robustness, the proposed framework uses a novel keypoint-based method to simultaneously predict hand regions and side labels, unlike existing methods that suffer from background color confusion caused by using segmentation or detection-based technology. Moreover, inspired by the biological structure of the human hand, we introduce two geometric constraints directly into the 3D coordinates prediction that further improves its performance in a weakly-supervised training. Experimental results show that our proposed framework not only performs robustly on unconstrained setting, but also outperforms the state-of-art methods on standard benchmark datasets.

%that can be used in a weakly-supervised setting to improve the performance of a 3D pose network by allowing training on external datasets without 3D annotations. Experimental results for the proposed hand detector and geometric constraints not only demonstrate the effectiveness of these methods for unconstrained 3D hand pose estimation, but also show they can achieve state-of-the-art performance on standard benchmark datasets.
\end{abstract}

% Note that keywords are not normally used for peer review papers.

% For peer review papers, you can put extra information on the cover
% page as needed:
% \ifCLASSOPTIONpeerreview
% \begin{center} \bfseries EDICS Category: 3-BBND \end{center}
% \fi
%
% For peerreview papers, this IEEEtran command inserts a page break and
% creates the second title. It will be ignored for other modes.
\IEEEpeerreviewmaketitle

\section{Introduction}
% The very first letter is a 2 line initial drop letter followed
% by the rest of the first word in caps.
% 
% form to use if the first word consists of a single letter:
% \IEEEPARstart{A}{demo} file is ....
% 
% form to use if you need the single drop letter followed by
% normal text (unknown if ever used by the IEEE):
% \IEEEPARstart{A}{}demo file is ....
% 
% Some journals put the first two words in caps:
% \IEEEPARstart{T}{his demo} file is ....
% 
% Here we have the typical use of a "T" for an initial drop letter
% and "HIS" in caps to complete the first word.
\IEEEPARstart{H}{and} pose estimation finds usage in wide array of applications like virtual or augmented reality, sign language recognition, gesture recognition, robotics, human-computer interface etc. Despite the large overlap in the set of problems and difficulties faced with human pose estimation, hand pose estimation has its own unique set of problems like lack of characteristic local features, heavy ambiguity, strong articulation and substantial self-occlusion making it a challenging problem to solve. Traditional methods \cite{quach2016depth,voxeltovoxel,ge20173d,deepprior++} that tackle these problems rely highly on depth or stereoscopic data, therefore have limited capacity in many real-world applications where images are captured by monocular cameras with no explicit depth information.

\begin{figure}
\centering
\includegraphics[width=1.0\columnwidth]{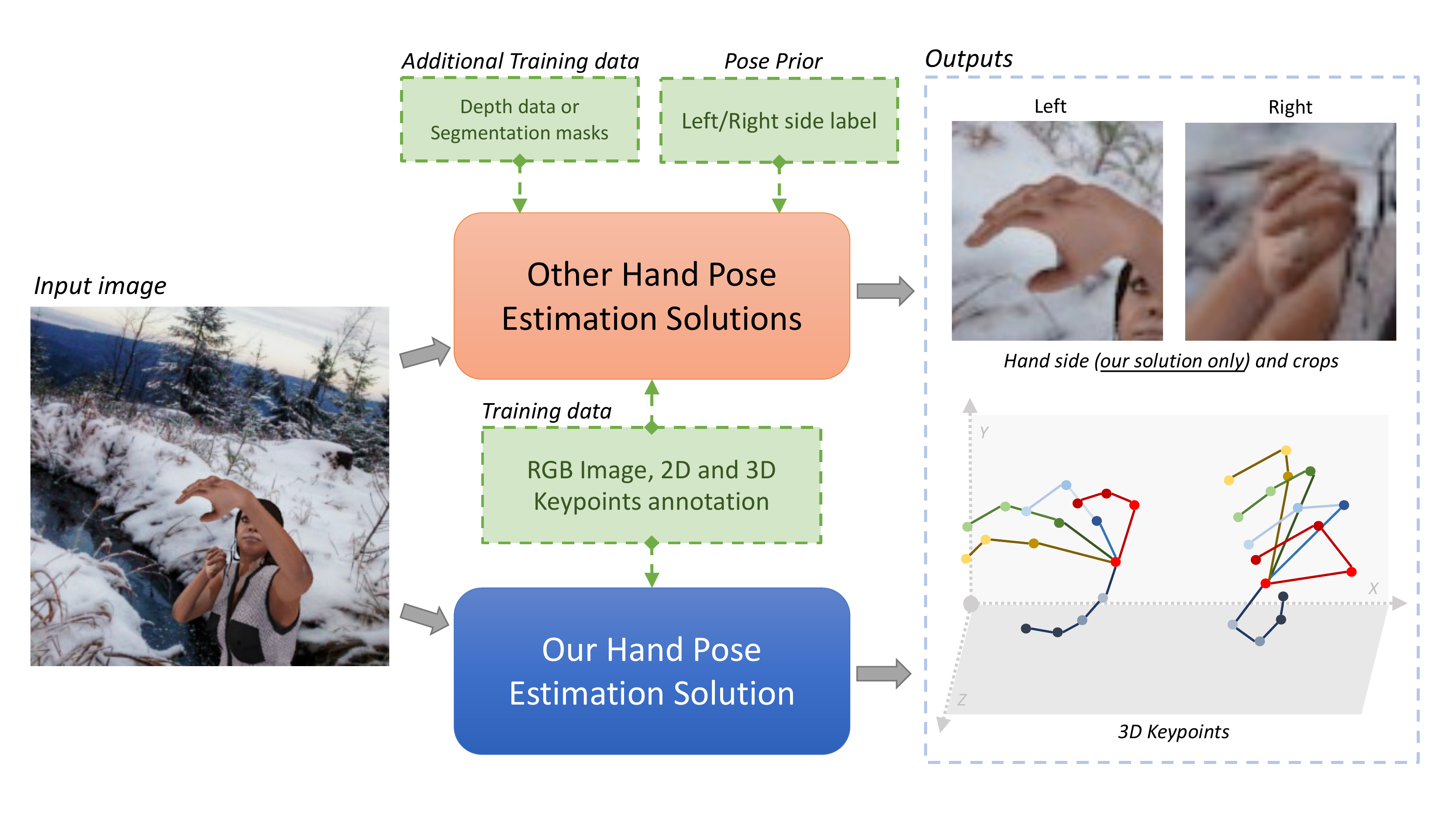}
\caption{\textit{A comparison of our proposed method with existing methods.} Compared to existing methods that require additional data annotations like depth maps, segmentation maps or hand side labels in addition to the usual RGB images and 2D and 3D keypoints annotations, our method only utilizes the latter three types of data while achieving superior performance in experiments. Again, our method is also able to predict hand side label in the same pipeline unlike any other existing methods.}
\label{fig:TitleImg}
\end{figure}

Recent interests in 3D hand pose estimation have been fostered by the prospects of monocular image-based methods. The related works including \cite{LearnToEstimate3DHandPoseFromSingleImage,CrossModelDeepVariationalHandPose,3DHandShapePoseInWild,ModelingConstraintsOfHandMotion,Weakly3DHandPose,GANHand3DTracking} have demonstrated that they can also achieve promising and effective results without the need of depth information from sensors. However, most of the design of these methods is based on the assumption that some prior information such as hand bounding boxes and left-right hand side label are already available. This makes them fail to run in an end-to-end manner on unconstrained image data. To this end, we propose an end-to-end framework that is capable of handling unconstrained monocular images for 3D hand pose estimation. The framework complements the current lack of 3D hand estimation system which is directly applicable to real-world environments.

To enable end-to-end processing for unconstrained images, a standard 3D pose system should have the ability to predict such essential prior information. This is due to the fact that the left-right hand side predicted label may be required in latter stages of 3D prediction, as in the architecture of \cite{LearnToEstimate3DHandPoseFromSingleImage}, as well as for real-world applications. Again, the quality of the extracted hand regions will largely affect the performance of subsequent hand pose estimation. For instance, if a low-quality hand region is used to generate the image crops, this will inadvertently result in an unrecoverable error during inference in the pose estimation network.

Recent works \cite{LearnToEstimate3DHandPoseFromSingleImage,GANHand3DTracking,RefinableNets} adopt an off-the-shelf segmentation network as an intermediate component in their 3D pose estimation framework to produce image crops from unconstrained images. Nonetheless, the performance of this segmentation-based approach for hand detection is unstable and unsatisfactory because it is usually confused by skin color objects. For example, if the target hand is located around the face or close to the other hand, the segmentation-based approach tends to produce an oversized cropped image, resulting in a smaller scale of the target hand. Again, its bounding box center will also offset from the hand center by a large margin leading to heavily cropped hand in the resultant image. In this work, we present a novel keypoint-based method for hand detection which performs robustly on unconstrained datasets. Unlike segmentation (pixel-wise) methods \cite{LearnToEstimate3DHandPoseFromSingleImage, RefinableNets, AnalysisOfHandSegInTheWild} that rely highly on color discrimination, our keypoint prediction-based method not only considers the visual appearance but also the spatial structure of the objects. Thus, by inference of hand regions from keypoint heatmaps, our method minimizes error caused by confusing color effect. At the same time, using only keypoints also avoids the need for expensive to acquire segmentation annotations. Besides, our proposed method can also simultaneously predict the left-right side of hands by adding two branches of keypoint prediction headers responsible for the left and right hand respectively.

In our proposed framework, we further improve the performance of 3D hand pose recovery by introducing two constraint functions inspired by the inherent anatomy of human hand. According to a study \cite{george1930human} on human hands, one general characteristic seen in them is that the middle finger is the longest while the thumb is the shortest, followed by the little finger. To model such a relative length relationships between fingers as a constrain function for 3D hand pose estimation, we introduce a length ratio term to enforce the predictions to follow the realistic relative length relations among fingers. Also, given the fact that each finger has its limited motion range, we introduce an angle range term to penalize the predictions containing unrealistic finger angles. In our proposed framework, these two constraints are directly applied into to 3D pose recovery module in a weakly-supervised setting during training, and both are used to reduce implausible hand pose results estimated by direct keypoint regression.

To validate the effectiveness of each proposed component and the overall framework, we conduct a series of experiments including hand and its side detection, 3D hand pose estimation with ground-truth hand crops, measurement of joint errors with the application of each constraint, and full pipeline 3D pose estimation.
To sum up, our main contributions are three-fold:
\begin{enumerate}
\item A novel keypoints-based method for hand detection that is more robust to some difficult situations such as confusing background and adjacent hands.
\item Two anatomy-based constraints for aiding 3D hand pose estimation network to improve its performance.
\item An end-to-end pipeline with state-of-the-art performance that can automatically process unconstrained images.

\end{enumerate}

\begin{figure*}[!t]
\centering
\includegraphics[width=1.0\textwidth]{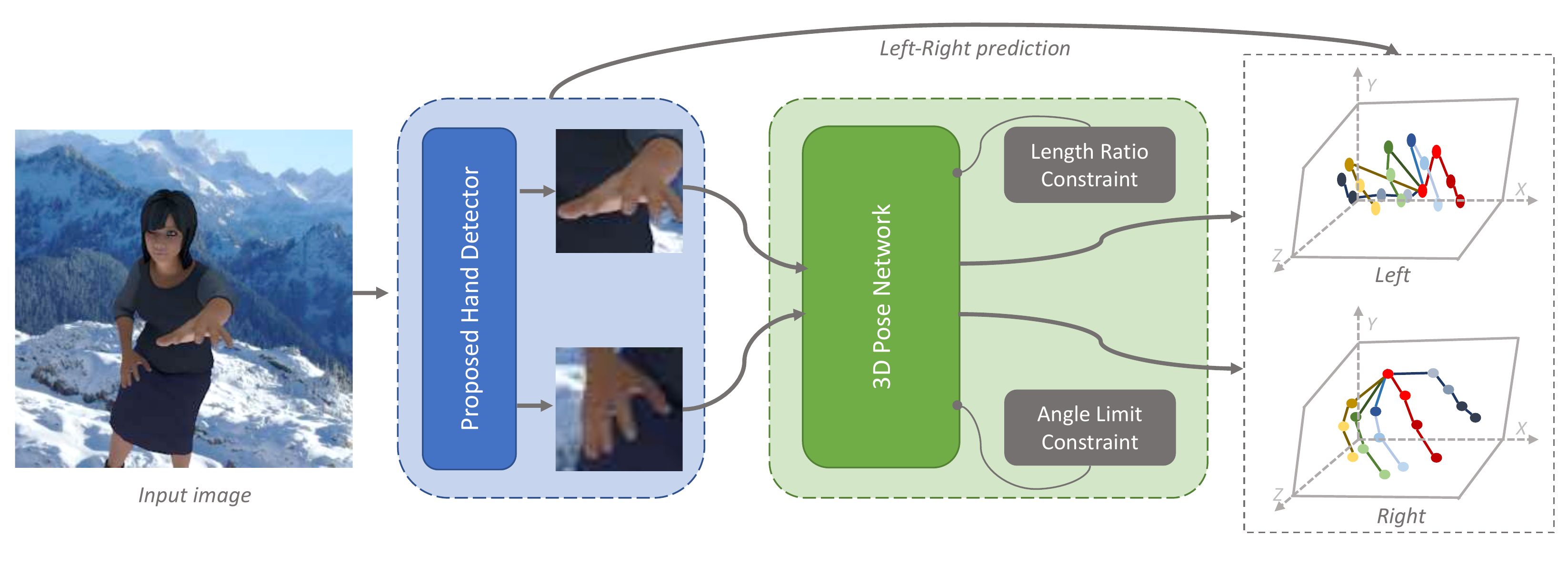}
\caption{\textit{An overview of the proposed full 3D hand pose estimation pipeline.} The input image is first processed by our hand detector which is capable of providing both left and right hand crop images along with their hand side label. These crop images are then processed by our \textit{3D Pose Network} trained with \textit{Relative Finger Bone Ratio} and \textit{Angle Range Constraints} to predict 3D hand coordinates.}
\label{fig:fullarch}
\end{figure*}

\section{Related Works}
\subsection{Traditional methods for Hand Pose Estimation}
Traditionally, 3D hand pose estimation solutions have made use of some form of depth information, either explicitly, by using depth maps or implicitly, by using multi-view images or hand mesh annotation. Among depth image-based solutions, DeepPrior++ \cite{deepprior++} is considered one of the important works that improves upon the original DeepPrior architecture \cite{deepprior} by adding ResNet layers, using data augmentation and developing better initial hand localization techniques. \cite{ge20173d}, is another depth-based method, which theorizes that image-based feature extraction using CNNs cannot effectively capture the complex 3D spatial information of the hand and, hence, proposes a 3D CNN architecture-based approach which works on a 3D volumetric representation instead. Similarly, \cite{malik2017simultaneous} focuses on solving the problem presented by varying hand shape and sizes using depth images. It considers kinematical properties and physical constraints of a human hand to accurately model its 3D keypoint predictions. Works such as \cite{ge20173d} and \cite{Weakly3DHandPose} make use of depth maps under an unsupervised setting to improve their network performance while other works such as \cite{choi2017robust} and \cite{hasson2019learning} make use of depth information to deal with occlusion problems caused when hands are holding objects.

Some methods make use of multi-view images which can provide an implicit form of depth information. \cite{MultiviewBootstrapping} is one such work for hand keypoints prediction that makes use of multi-view images from a specifically built multiple camera setup. In this method, using at least two images that have correctly predicted keypoints with a weak detector, a re-projection is calculated for each keypoint in other unannotated views. The triangulated points are used to select N top performers which are then used to retrain the weak detector in an iterative manner. While this method performs very well with any in-the-wild images for hand 2D keypoints prediction, 3D keypoint prediction is only available inside their camera setup. Another work by \cite{wang20116d} makes use of a two-camera setup in a fixed setting to determine hand pose for manipulating CAD designs. Each detected hand has two segmented views which are used to look up the closest matching pose in a database. This method is limited not only in its use case scenario but also due to its assumption that all kinds of complex hand articulations can be summarized in a database. Finally, other works such as \cite{ballan2012motion}, \cite{sridhar2013interactive} and \cite{sridhar2014real} make use of up to 8 calibrated cameras in a multi-camera setup with an additional depth sensor in case of the latter two. Compared to collecting depth maps, multi-view image collection requires a lot more complicated setup.

More recently, with the invention of Graph Convolution Networks (GCNs) \cite{graphconv1, graphconv2}, hand modelling with mesh prediction-based approaches using graph convolutional networks have also gained some popularity. Unfortunately, with the hand pose estimation domain already suffering from limited datasets for annotations besides depth maps, these methods must find clever methods to adapt existing annotations to synthetically generate mesh  while also requiring the creation of a separate dataset. \cite{ge20193d} requires both mesh and depth map annotation to jointly train for 3D hand mesh and 3D keypoints prediction. Similarly, \cite{kulon2019single} tries to encode images into a latent space for a non-linear representation of hand mesh. None of these works tackle the problem of creating a hand crop or predicting hand side.

Even though the cost of hardware used to capture depth data, generate sets of multi-view images of a hand and artificially synthesize hand mesh annotation required by aforementioned solutions have come down considerably over the years, the practicality of using these types of hardware in a real-world scenario remains limited. Rather, a typical real-world usage scenario would consist of a color image of a person as an input taken with a single RGB camera. A complete hand pose estimator is expected to not only parse hand locations but also predict 3D keypoints from this single image which lacks any form of depth data. This makes hand pose estimation on monocular RGB image an ill-posed problem. \cite{LearnToEstimate3DHandPoseFromSingleImage} is among the first works that tackles the problem of hand pose estimation on monocular images using deep learning formulation. This work provides a complete pipeline with a hand detector and hand pose estimator. Their \textit{SegNet}-based hand detector, unfortunately, is only capable of detecting the largest hand in an image. Again, as their hand detector is unable to provide hand side label required by their hand pose estimation part of the network, this information is assumed to be a prior. \cite{GANHand3DTracking}, on the other hand, jointly tackles the problem of limited dataset in the domain with monocular RGB-only based hand tracking by making use of their GAN-based synthetic-to-real hand image translation network. Similarly, \cite{CrossModelDeepVariationalHandPose} also tries to remedy the problem of limited RGB dataset by making use of all the different types of annotation available to learn a cross-model latent space embedding. An input RGB image is then used to predict 3D keypoints from this space. Finally, works like \cite{Weakly3DHandPose, 3DHandShapePoseInWild} and \cite{ge20193d} try to jointly learn tasks of either depth map prediction or hand shape modelling to improve performance of a network.

\subsection{Human Pose Estimation from Single Monocular Images}
As the domains of hand pose and human pose estimation face similar set of difficulties, we review some of the important works on human pose estimation that operate on single monocular images. The work \cite{moreno20173d} theorizes that by making use of Euclidean Distance Matrices (EDM), it is possible to derive more precise estimates of 3D coordinates from 2D predictions. They claim that their method counters difficulties of missing observations and depth-scale variance. These problems also occur very frequently in hand pose estimation domain. So far, no work has attempted to use distance matrix regression in hand pose estimation. The paper \cite{tekin2017learning} tries to combine the currently two popular streams of 3D coordinate predictions namely – regressing directly from image and using predicted 2D joint locations to infer 3D joint coordinates. The fusion scheme proposed by them is fully trainable removing the need to choose between early and late fusion. The authors show that this kind of model is very effective in exploiting cues in monocular RGB images to infer 3D joint locations. \cite{coarsetofine} proposes a fine discretization of 3D space around the human body using voxels. The pose estimation task is converted to predict voxel likelihoods for each joint. The authors claim that this method is superior to traditional approaches of direct regression of 2D joint coordinates. In addition, their architecture employs a coarse-to-fine scheme for prediction which they claim further improves performance. A similar voxel-based approach has been proposed for hand pose estimation by \cite{voxeltovoxel}. Two important works \cite{Learn3DFromStructureAndMotion} and \cite{Towards3DHumanPoseInWild} introduce a weakly-supervised setting for making use of known constraints of human body with unlabeled training images from the wild. Our proposed method for hand pose estimation is inspired from these works to exploit biological geometrical constraints of human hands.

\begin{figure*}[!t]
\centering
\includegraphics[width=1.0\textwidth]{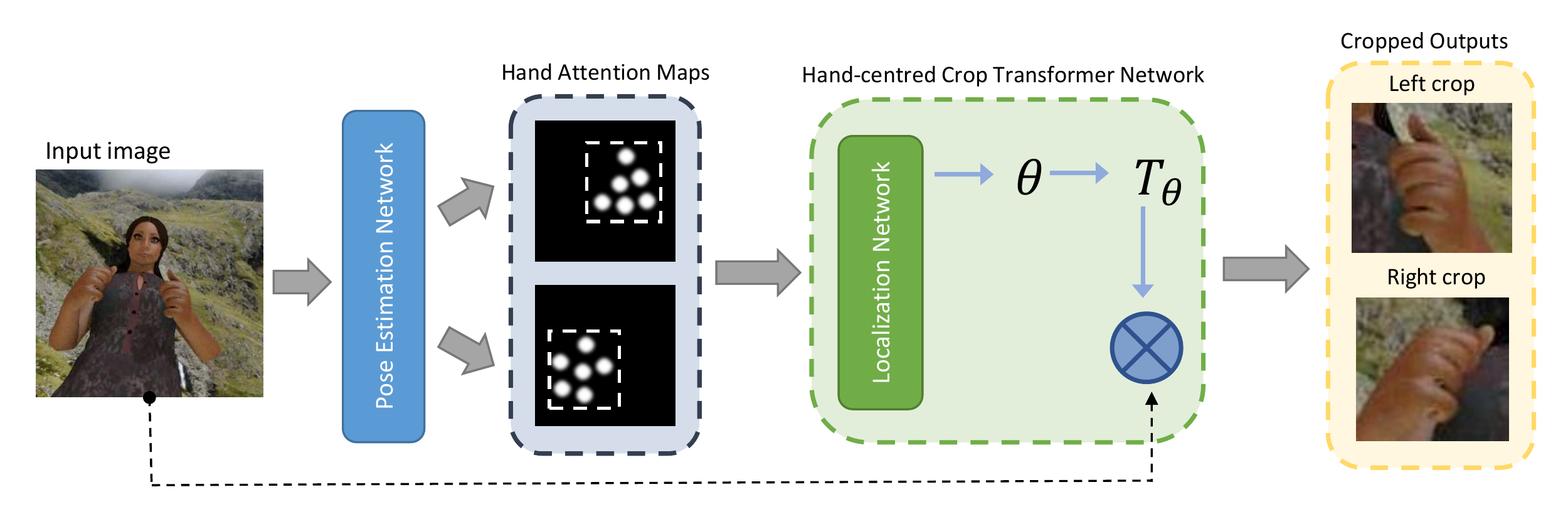}
\caption{\textit{The components of our proposed Keypoint-based Hand Detector.} The \textit{Left-right Hand Keypoint Attention Network} predicts \textit{Hand Attention Maps} for keypoints in each hand. The maps are processed by \textit{Localization Network} in \textit{Hand-centered Crop Transformer} to generate transformation parameters. These parameters are then applied to the original input image to create a crop image for each hand given that a specific hand is visible in the input image.}
\label{fig:handdetectarch}
\end{figure*}

\subsection{Hand Detector for Hand Pose Estimation}
Most of the works in hand pose estimation assume that the input image to their network are an already well-cropped hand and so, consider hand detection part of the pipeline to be out-of-scope in their works. Some works like \cite{LearnToEstimate3DHandPoseFromSingleImage} have proposed a segmentation-based hand detector that has been used either without modification like in \cite{GANHand3DTracking} or with some modification, designed only to reduce complexity without improving much performance, as in \cite{RefinableNets}. Despite the negative effect on the performance noted by \cite{LearnToEstimate3DHandPoseFromSingleImage} due to their hand-detector on the overall pose detection pipeline, there are very limited literature studying hand detectors. \cite{AnalysisOfHandSegInTheWild} and \cite{bambach2015lending} study segmentation-based hand detection but their setting is for an egocentric view. \cite{InWild}, on the other hand, makes use of \textit{YOLOv2} as a hand detector. They use a rudimentary method of determining hand side by detecting whether a detected hand is on the left or right side of the person’s head. In real world use case, when a person crosses his/her hand or when the head of the person is not visible in the image, this method can easily fail. Similarly, popular pose estimation frameworks such as \textit{OpenPose} assume that hand side for their implementation of hand pose estimator can be determined from adjoining arm keypoint prediction label. Running a full human pose detection model just to determine arm keypoint will not always be practical as it requires lot of redundant calculations for unused body keypoints. The work \cite{chen2016deep} proposes a novel method of combining hand detection and estimation stages into a single pipeline but the proposed solution works only for depth map input.

\subsection{Geometry Constraints for Pose Estimation}
Human body parts have different kinematical and structural constraints \cite{ren2005recovering} imposed by biological evolution. \cite{Towards3DHumanPoseInWild} exploits the fact that fixed ratios exists between bone lengths of different body parts to propose a geometric constraint that is used as an regularizer for their depth prediction when ground truth for depth data is not available. Similarly, \cite{Learn3DFromStructureAndMotion} exploits the fact that body joints can only bend in a fixed angle range and that left and right body parts are symmetrical. Using these facts, two sets of losses are proposed that they use to improve their model with data that lacks full 3D annotation. Similar joint angle constraint is also used by \cite{GANHand3DTracking} for their kinematical skeleton fitting hand model. \cite{ModelingConstraintsOfHandMotion} proposes a learning based approach to model constraints in a configuration space. This work focuses on learning local finger motions of different sets of finger digits rather than the global hand motion.

\begin{figure*}[!t]
\centering
\includegraphics[width=1.0\textwidth]{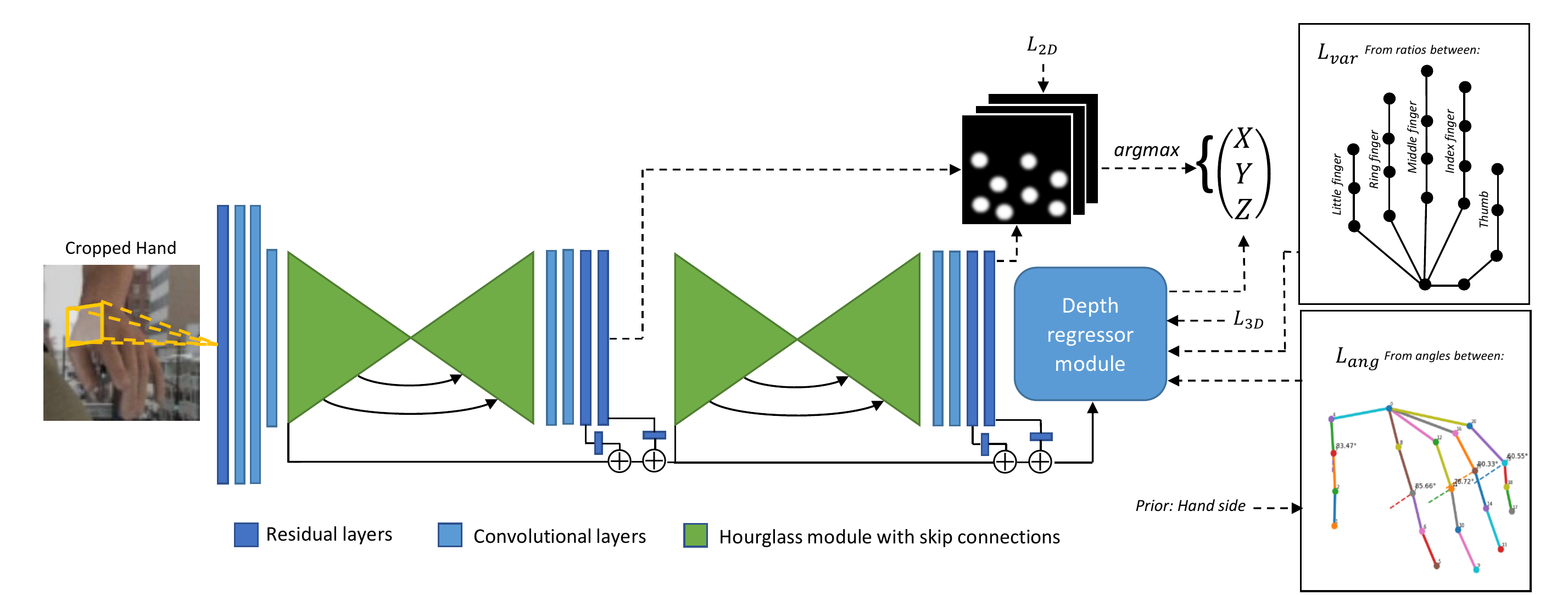}
\caption{\textit{3D hand pose estimation part of the pipeline trained using proposed constraints to predict 3D hand coordinates.} Two geometric constraint-based losses help \textit{Depth regressor module} to learn the 3D structure of the human hand better.}
\label{fig:handposearch}
\end{figure*}

\section{Framework}
Our proposed framework consists of a keypoint-based hand detection and a 3D hand pose estimation network with two geometric constraints. As illustrated in Fig. \ref{fig:fullarch}, the first component takes an unconstrained image as input and automatically produces hand-centered image crops that are directly and sequentially fed to the second component to estimate the 3D poses. By combining the predicted results from the first component, the proposed framework can estimate 3D hand poses with left-right labels directly from an unconstrained monocular image.

We first, briefly introduce the notation. Let $\textbf{\textit{D}} = \{(I_{i},o_{i})\}_{i=1}^{N}$, where $N$ is the number of sample. Each sample $(I,o)$ includes an image $I$ and the ground-truth joint coordinates $o=\{o^{2d},o^{3d}\}$, where $o^{2d} = \{(x^i,y^i)\}_{i=1}^{K}$ and $o^{3d} = \{(x^i,y^i,z^i)\}_{i=1}^{K}$ denotes the $K$ number of 2D and 3D ground-truth joint coordinates respectively. We also denote $S=\{S^j\}_{j=1}^{K}$ as the ground-truth heatmaps for one hand, where $S^j$ refers to the heatmap of joint $j$ and is generated from a Gaussian centered at $(x^i,y^i)$. Similarly, $\tilde{S}$ is the predicted heatmaps, one for each keypoint, by the CNN network.

\subsection{Keypoint-based Hand Detection}
The main idea of the keypoint-based detector is to locate the hand regions based on the keypoint attention maps and automatically produce centered and resized image crops by a \textit{Hand-centered Crop Transformer}. Also, to obtain the left-right side information, we explicitly differentiate left- and right-hand predictions by adding two separate prediction headers on top of a shared \textit{Conv} feature map. As shown in Fig. \ref{fig:handdetectarch}, the proposed detector consists of: 1) a fully convolutional network that simultaneously predicts two separate keypoint attention maps for each hand side and, 2) a transformer network that automatically generates hand-centered image crops.

\hfill
\break
\noindent
\textbf{Left-right Hand Keypoint Attention Network}
We first, learn a 2D keypoint prediction network on the unconstrained data to obtain hand attention regions. We follow the standard pipeline \cite{CPM} for 2D pose estimation but with a slight difference that we explicitly predict two sets of keypoints for both hands. We formulate this problem as a heatmap regression problem. Given $LS^j$ and $RS^j$ for the left-hand and right-hand ground-truth heatmap for hand joint $j$ respectively, and similarly $\tilde{LS}^j$ and $\tilde{RS}^j$ for the predicted heatmap respectively, the loss for training hand attention network is defined as follows:
\begin{equation}
    \mathcal{L}_{att} = \frac{1}{N}\sum_{i=1}^{N}\sum_{j=1}^{K}\left(\|{LS}_i^j - \tilde{LS}_i^j\|_{2}^{2}+\|{RS}_i^j - \tilde{RS}_i^j\|_{2}^{2}\right).
    \label{eq:1}
\end{equation}

In this module, we adopt an existing pose estimation network as the backbone network structure, such as \textit{Convolution Pose Machine} (CPM) \cite{CPM} and \textit{Stacked Hourglass Networks} (HG) \cite{stackedhourglass}. By training with the loss in Eq. (\ref{eq:1}), we then can learn to predict coarse hand heatmaps $\{\tilde{LS}^j,\tilde{RS}^j\}$ from an input image. To determine the hand side, we use a simple thresholding method on heatmaps. Our investigation showed that when a hand was not in an image, the Gaussian peaks $\mathcal{P}=\max(\frac{1}{K}\sum_{j=1}^{K}{S}^j)$ in heatmaps for that hand side were very low and sparsely spread. This allowed us to determine if a certain hand was missing.

\begin{figure*}[!t]
\centering
\includegraphics[width=1.0\textwidth]{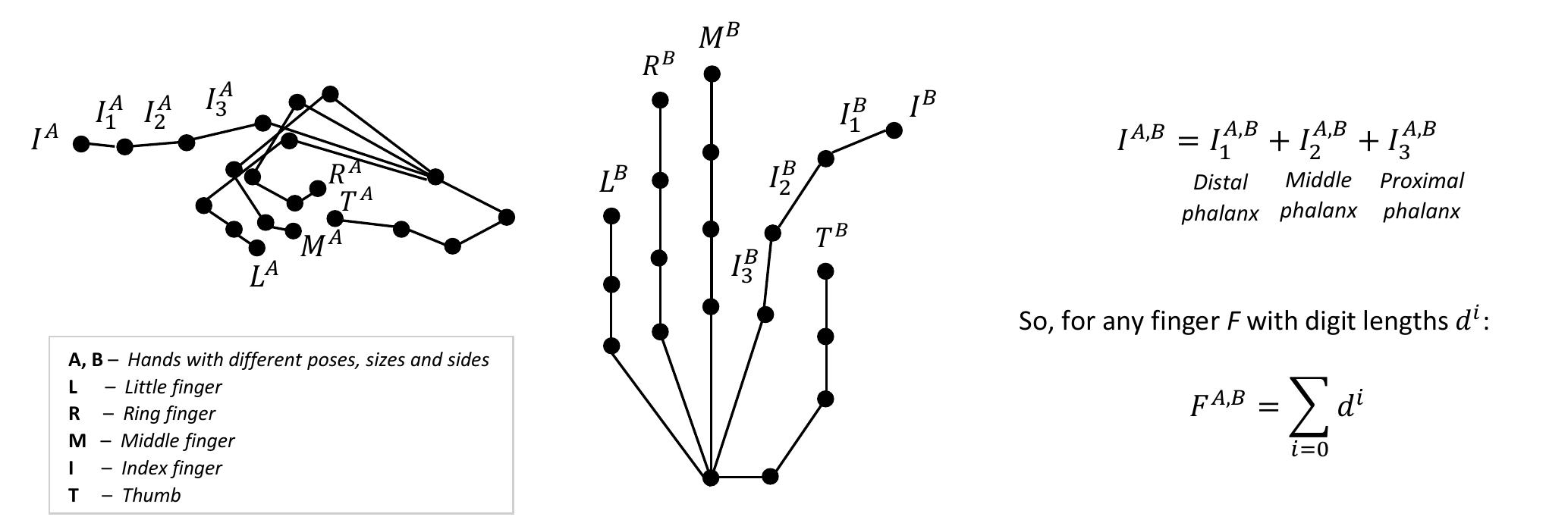}
\caption{\textit{Method to calculate bone lengths from digit lengths.} Biological formation of a human hand determines the relative sizes of each finger to every other finger. Using 3D coordinates of different digits in a finger, we can calculate the length of a finger by summing the length of its digits and hence calculate the average ratio of a finger to every other finger in a training dataset. Using these average ratios with \textit{Relative Finger Bone Ratio Loss}, we provide a constraint to use in weakly-supervised setting during training.}
\label{fig:HandPose_BoneLengthConstraints}
\end{figure*}

\hfill
\break
\noindent
\textbf{Hand-centered Crop Transformer}
To get an appropriate hand crop from the input image, one simple way is to use a box coordinates calculated from the the predicted heatmaps with predefined cropping parameters for post-processing an input image. However, this method is fully manual and the thresholds need to be carefully selected to determine the box coordinates. Also, when the pose estimation network incorrectly predicts some keypoints, this method might encounter an unrecoverable error for cropping out the hand region. Thus, we propose a \textit{Hand-centered Crop Transformer} that learns to automatically generate an appropriate hand region based on the predicted heatmaps and allows passing the error signal back to \textit{Pose Estimation Network} when bad transform parameters are predicted. We derive the inspiration of this component from works like \textit{Spatial Transformer Networks} (STN) \cite{STN} and \textit{Attention Proposal Networks} (APN) \cite{LookCloser}. These networks have been trained in various literature to produce corrected versions and/or finer zoomed crops of their input to aid later stages. Unfortunately, they cannot be employed directly for our case because these transformer networks rely on ranking or hinge loss to iteratively learn how to produce a good set of affine transformation parameters. Hence, we propose a cropping loss for learning this component. Given the perfect square box coordinates $b=[b_{x1},b_{y1},b_{x2},b_{y2}]$ calculated from the ground-truth heatmaps $S$, we expect the localization network $f_\mathcal{W}(\cdot)$ parameterized by $\mathcal{W}$ to take the predicted heatmaps $\tilde{S}$ as input and predict a transformation parameter matrix $\theta$, so that an affine transformation function $T_{\theta}(\cdot)$ can transform the box coordinates $b$ to match coordinates $\tilde{b}=[0,0,w-1,h-1]$, where $w$ and $h$ are the width and height of the output heatmaps respectively. Hence, the cropping loss can be expressed as:
\begin{equation}
    \mathcal{L}_{crop} = \|T_{f_W(\tilde{S})}(b) - \tilde{b}\|_{2}^{2}.
    \label{eq:2}
\end{equation}

For training the hand detector, we first train the pose network with the loss $\mathcal{L}_{att}$ and transformer network with the loss $\mathcal{L}_{crop}$ separately, then fine-tune both networks jointly with loss  $\mathcal{L}_{det} = \mathcal{L}_{att} + \lambda \mathcal{L}_{crop}$, where $\lambda$ is a hyper parameter to be chosen. In the testing stage, 
 we apply the affine transformation with the predicted parameters to obtain the hand crops for the 3D hand pose estimation.

\subsection{Anatomy-aware Net for 3D Hand Pose Estimation}
Our proposed framework for 3D hand pose estimation, as shown in Fig. \ref{fig:handposearch}, consists of a baseline pose network architecture as in \cite{Towards3DHumanPoseInWild} and two anatomy-inspired losses that aid into the network learning.

\hfill
\break
\noindent
\textbf{Baseline Pose Network}: The baseline pose network consists of a 2D regression module, a depth regression module and two geometric constraints. Specifically, the 2D regression module is supervised by 2D heatmap regression loss, which is defined as:
\begin{equation}
    \mathcal{L}_{2D} = \frac{1}{N}\sum_{i=1}^{N}\sum_{j=1}^{K}\|{S}_i^j - \tilde{S}_i^j\|_{2}^{2}.
    \label{eq:3}
\end{equation}

The depth regression module is used to learn depth when there is input from 3D data. But when 3D data is not available, we penalize its 3D prediction with loss from geometric constraints. Hence, the learning loss is expressed as:
\begin{equation}
     \mathcal{L}_{depth} = \begin{cases}
    \lambda_{reg}\mathcal{L}_{reg},                 & if  I \in D_{3D} \\
    \lambda_{geo}\mathcal{L}_{geo}, & \text{otherwise.},
     \label{eq:4}
    \end{cases}
\end{equation}

where $I \in D_{3d}$ indicates if input come from the 3D training data, $\lambda$s are hyper parameters for adjusting the contribution of each loss, $\mathcal{L}_{geo}$ encompasses the body anatomy inspired geometric constraints, and $\mathcal{L}_{reg} = \frac{1}{N}\sum_{i=1}^{N}\|{z}_i - \tilde{z}_i\|_{2}^{2}$. Thus, the overall loss can be defined as:
\begin{equation}
    \mathcal{L}_{base} = \mathcal{L}_{2D} + \mathcal{L}_{depth}.
    \label{eq:5}
\end{equation}

To adapt this weakly-supervised method for human pose estimation to the fully-supervised task of hand pose estimation, we first modify the depth regression module by removing the term when input is from 2D data. Also, recent work \cite{Weakly3DHandPose} shows that \textit{smooth L1 loss} performs better than \textit{L2 loss} in hand pose estimation, therefore we use \textit{smooth L1 loss} to measure error of predicted depth. The modified depth loss function from Eq. (\ref{eq:4}) is now defined as:
\begin{equation}
     \mathcal{L}^{L1}_{depth}(e_i) = \begin{cases}
    \frac{1}{2}{e_i^2}               & \text{for } |e_i| \le \delta, \\
    \delta (|e_i| - \frac{1}{2}\delta), & \text{otherwise.},
    \end{cases}
\end{equation}

where $e_i=\|{z}_i - \tilde{z}_i\|_{2}^{2}$ and $\delta$ is a predefined threshold.

\begin{figure*}[h!]
\centering
\subfloat[]
{
    \includegraphics[width=0.36\linewidth]{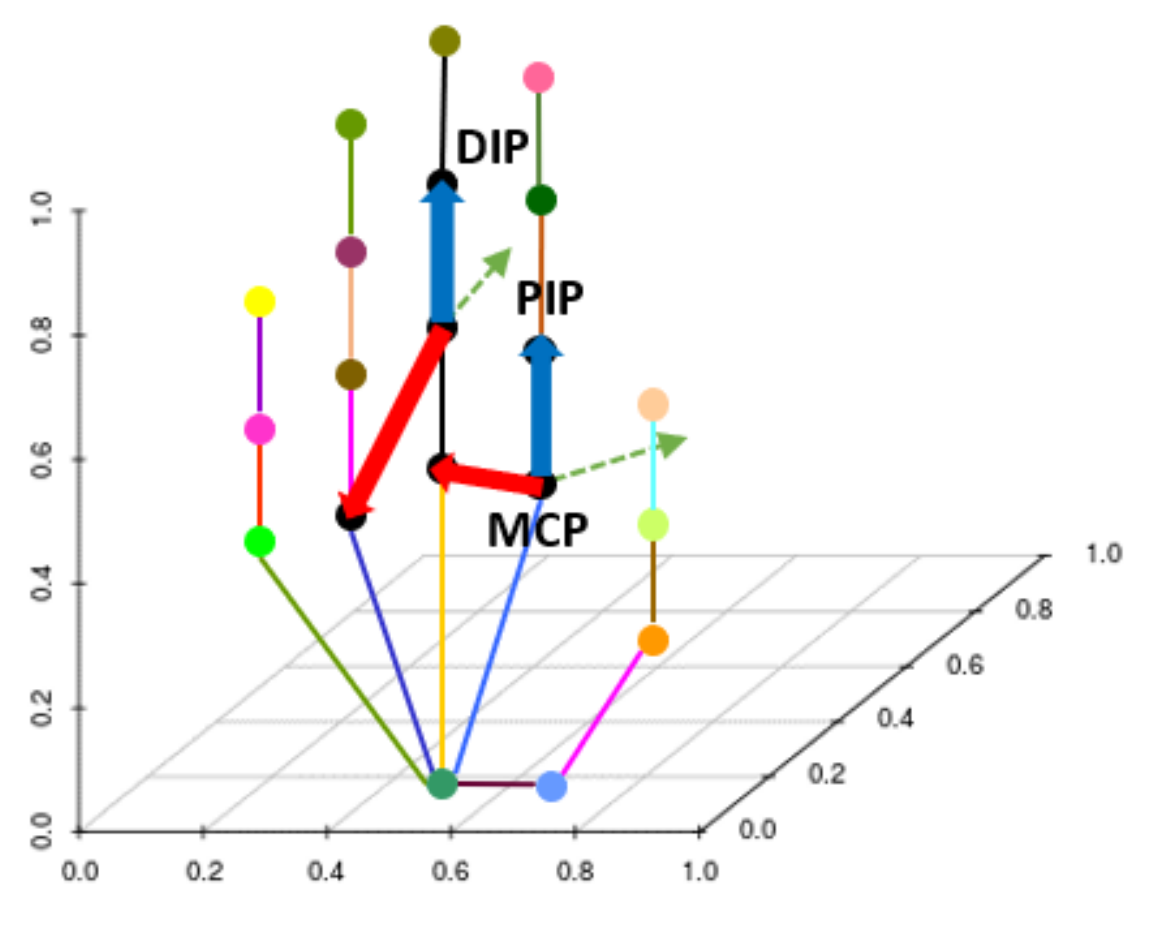}
}
\subfloat[]
{
    \includegraphics[width=0.32\linewidth]{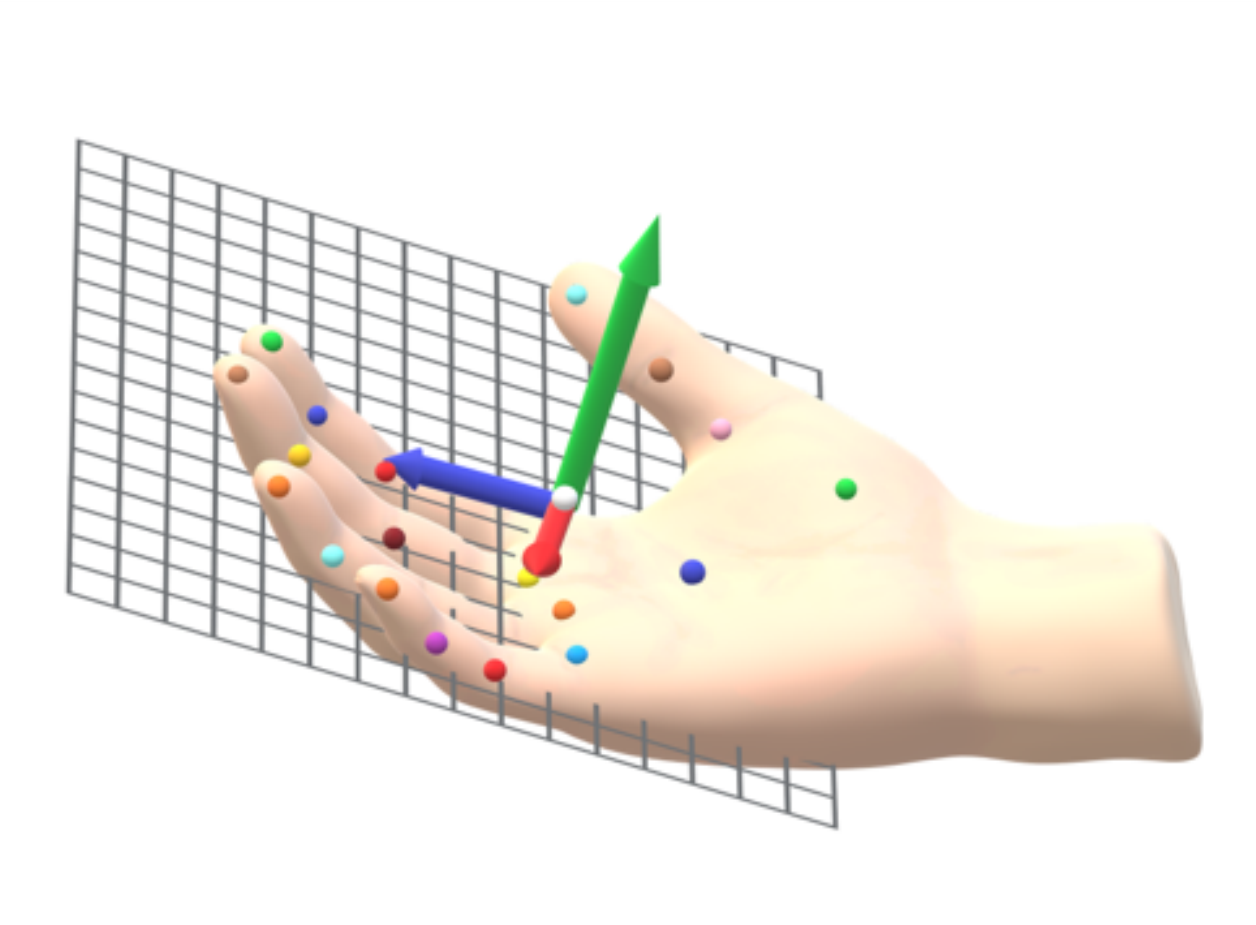}
}
\subfloat[]
{
    \includegraphics[width=0.31\linewidth]{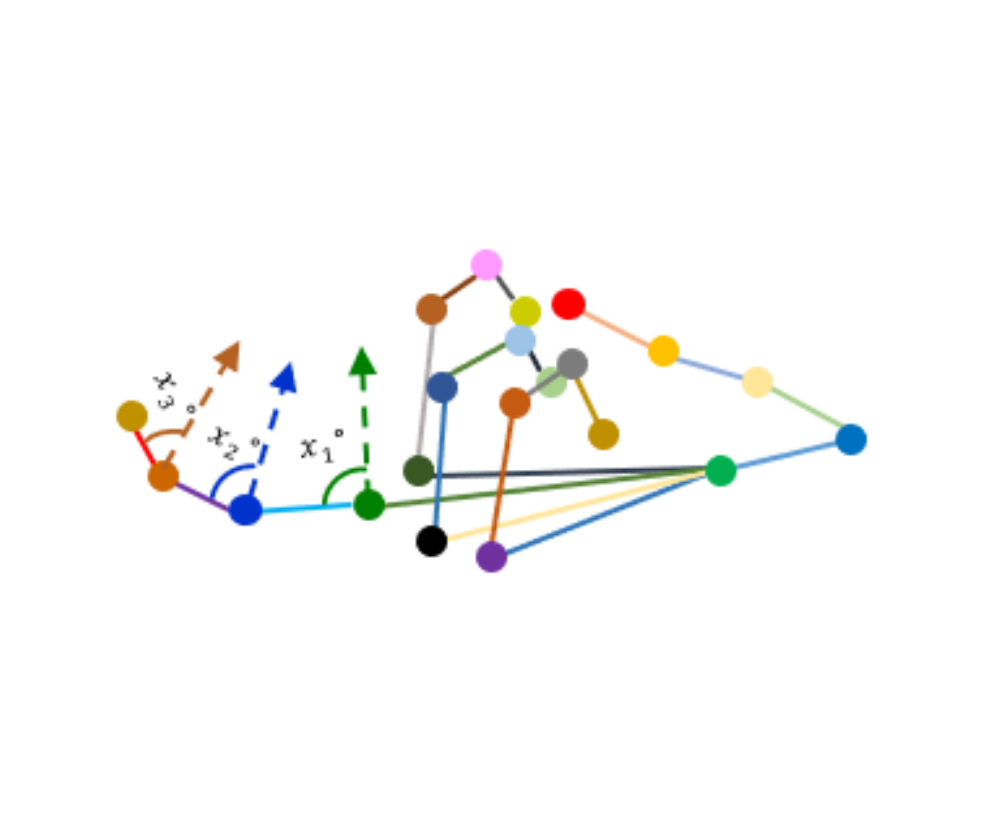}
}
\bigskip
\caption{The image (a) shows the two vectors (denoted by red and blue arrows) used to calculate the normal vector (green arrow) pointing towards the inside of the hand. This is necessary to always calculate the interior digit joint angles that make up the digit bends. The image (b) shows these normal vector more clearly on an actual hand model. Finally, the image (c) shows the interior angles that are calculated from the normal vectors on to their adjoining digits.}
\label{fig:AngleLoss}
\end{figure*}

%\begin{equation}
%    \mathcal{L}_{depth} = \frac{1}{N}\sum_{i=1}^{N}\sum_{j=1}^{K}\|{z}_i^j - \tilde{z}_i^j\|_{2}^{2}.
%    \label{eq:3}
%\end{equation}
%To enable weakly supervise leaning with 2D data, the geometry constraint $\mathcal{L}_{geo}$ is used to penalize 3D prediction with unreasonable bone length. Thus, the overall loss for learning the baseline pose network can be expressed as 
%The baseline pose network is proposed in [] for learning with both 2D and 3D training data. The network structure consists of a stacked hourglass module for feature learning, a 2D heatmap regression module for estimating 2D pose, a depth regression module for 3D pose recovery from 2D pose and a geometry constraint to penalize 3D prediction with unreasonable bone length. 

%To enable weakly supervise leaning with 2D data, [] also introduce a geometry constraint $\mathcal{L}_{geo}$ into the objective function of 3D pose estimation. Specifically, the 2D heatmap loss to learn learn the 2D heatmap and de re

\hfill
\break
\noindent
\textbf{Hand Anatomy Inspired Constraints}:
The human hand is a highly versatile and articulated body part. Hence, trying to model a hand using only image features is an ill-posed problem. But given the biological make up of hands, we can make use of its geometric and kinematical constraints to alleviate this problem. Therefore, we introduce two constraints related to hand anatomy to enforce a reasonable hand configuration on the pose network predictions.

%If we simplify hand skeleton to stick structure as in Fig. \ref{fig:HandPose_BoneLengthConstraints}, we can view each of the bones as a kinematical chain starting from either palm center or wrist. In our work, we use palm center. This hand model shows us that the four fingers have four degrees of freedom. The metacarpo-phalangeal joint have two degrees of freedom while the distal minterphalangeal and procimal interphalangel joint have one each \cite{ModelingConstraintsOfHandMotion}. The two degrees of for the former are possible due to flexion and abduction motions. The thumb has a very different structure to the other fingers and consists of five degrees of freedom.

%To make use of this hand configuration, we propose two main types of constraints to use as loss.

\subsubsection{Relative Finger Bone Ratio Loss $\mathcal{L}_{fr}$}
Although specific finger lengths among hands vary with different factors, such as genders and ethnic groups, their relative ratios remain relatively the same. For example, in terms of the ratio between the index and ring finger, the standard deviation for a selected group of people studied was found to be $0.032$ \cite{dalton2014self}. Thus, we introduce a relative finger ratio constraint to prevent the pose network from predicting results with unrealistic finger ratio. To this end, we try to reduce the variance across predictions by pulling the average of all bone length ratios between pairs of each finger in a hand towards the average calculated from all of training set.

To calculate the average of all bone length, we first need to calculate the length of each finger. Fig. \ref{fig:HandPose_BoneLengthConstraints} shows that we can calculate the length of a finger by summation of the length of each digit in 3D space. Hence, length of a finger $I_n$ is given by:
\begin{equation}
    I_n = \sum_{d}^{D} I^n_d,
\end{equation}

where $I^n_d$ is the length of d\textsuperscript{th} digit for n\textsuperscript{th} finger.

Then, we can calculate the average relative finger bone length ratio for a hand $\bar{R}$ as follows:

\begin{equation}
    \bar{R} = \frac{1}{^NC_2} \times \sum_{n}^{N} ((\sum_{m=n+1}^{N} I_m)/I_n),
    \label{eq:8}
\end{equation}

where $N$ is total number of fingers on a hand and $I_n$ is the length of a finger.

Finally, our finger bone ratio loss function $L_{fr}$ can be expressed as:
\begin{equation}
    \mathcal{L}_{fr}(\hat{R_i}, \bar{R}) = ||\hat{R_i} - \bar{R}||_2^2,
\end{equation}

where $\hat{R_i}$ is the prediction for one hand while $\bar{R}$ is computed from training set.

\subsubsection{Angle Range Loss $\mathcal{L}_{ar}$}
The angle between two digits in a finger is always inside a certain range due to constraint imposed by hand anatomy \cite{ren2005recovering}. Therefore, we further introduce an angle range loss that penalizes predictions where digits are bent beyond their acceptable ranges in relation to the immediate digit they are attached to. There are two finger digit motions that are important for our consideration - namely flexion and abduction.

In flexion motion, a digit can be flexed inwards towards the inside of the hand (i.e. towards palm). Depending on the position of a digit, there are different limitation of angles that they can flex at. In our work, we consider two flexion angle ranges - one of first digit (proximal phalanx) relative to middle digit (middle phalanx) and other of middle digit to last digit (distal phalanx). Similarly, we also used angle ranges by abduction motion which is the motion of first digit (proximal phalanx) in relation to other first digits of adjacent fingers.

In order to combine all these constraints and use them in a loss for our training, the method for angle calculation needs to be differentiable. We can easily calculate the angle between two 3D vectors using dot product. Dot products are differentiable making them suitable for use to calculate angle loss. Unfortunately, they only calculate acute and obtuse angles between two 3D vectors. This means if a prediction contains a flex of more than $180^{\circ}$ (i.e. reflex angle), which could be outside valid range, a loss using dot product will not penalize the network as expected.

Hence, we propose a new method of calculating angle between digits which makes sure that the calculated angle is always in the direction of the inside of the hand. In our method, we first calculate a normal vector between two adjacent digits. The direction of this vector must be towards the inside of the hand. To calculate such a normal vector between the first and middle digit of a finger, we use two Metacarpophalangeal (MCP) joints and one Proximal Interphalangeal (PIP) joint. By taking 3D points of these joints in a certain order to compute two vectors, we can make sure that the normal vector generated will met the directional requirement. Similarly, for middle and last joint, we make use of one Metacarpophalangeal joint, one Proximal Interphalangeal (PIP) joint and one Distal Interphalangeal (DIP) joint. This whole process is shown in Fig. \ref{fig:AngleLoss}(a).

When calculating normal vector as discussed above, we start from thumb progressing towards little finger. For little finger, we do not have a MCP joint in the direction required to calculate the proper normal vector. So, we use the MCP joint of the index finger and change the order of 3D points to get the correct direction for normal vector. Similarly, depending on hand side, we will need to change the order of 3D joint points used to create a normal vector. Alternatively, we could calculate in the same order for both hand sides and invert the direction of normal vector to correct it for one hand side by multiplying its unit vector with $-1$.

\begin{figure*}[!t]
\centering
\includegraphics[width=1.0\textwidth]{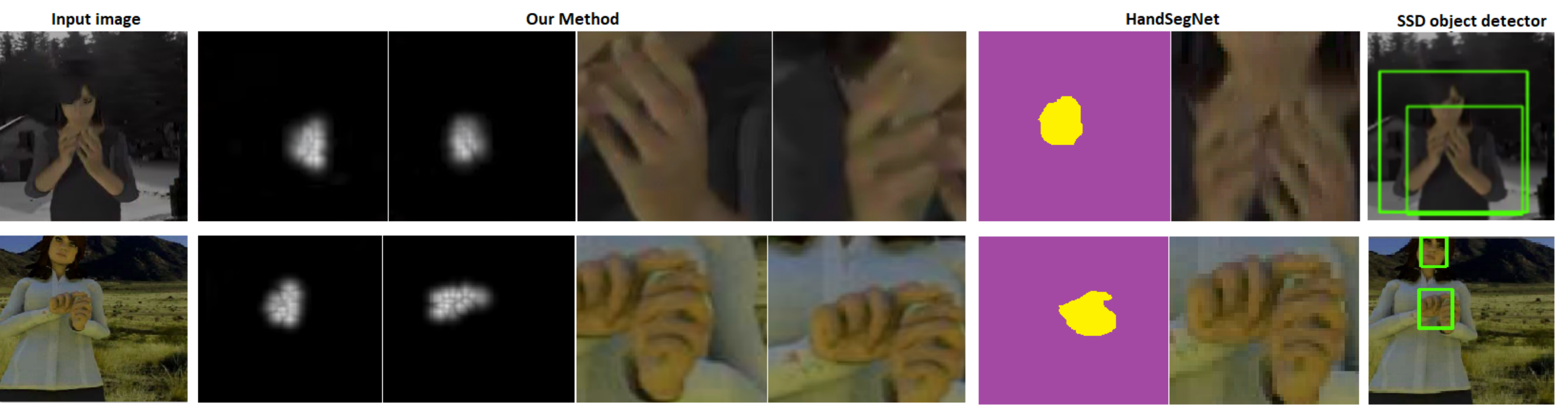}
\caption{Illustration of how our proposed method, \textit{HandSegNet} and \textit{SSD object detector} perform on inputs where the hands are close to each other or close to a person's face in the input image. Results show that our method not only performs better but also tries to bring the hand to the center of the crop. This may aid finer keypoints estimation. \textit{HandSegNet} runs into problems when two hands are close to each other while object detector fails completely in images with less illumination, confuses face with hands and produces a combined crop for both hands when they are close to each other.}
\label{fig:HandDetect_Samples}
\end{figure*}

%\noindent
%\textbf{Learning 3D Pose Network with Anatomy Inspired Constraints}

To incorporate these two constraints into the learning of 3D hand pose estimation, the overall loss can be expressed as:
\begin{equation}
    \mathcal{L} = \mathcal{L}_{2D} + \mathcal{L}^{L1}_{depth} + \beta_{fr}\mathcal{L}_{fr} + \beta_{ar}\mathcal{L}_{ar},
    \label{eq:6}
\end{equation}
where $\beta$s are hyper-parameters to be chosen.

When using the constraints in a weakly-supervised setting, we follow the same stage-based training approach as in \cite{Towards3DHumanPoseInWild} by first training with available 3D annotation using:
\begin{equation}
    \mathcal{L}^{stage1} = \mathcal{L}_{2D} + \mathcal{L}^{L1}_{depth}.
\end{equation}

Followed by joint training as the second stage with:
\begin{equation}
    \mathcal{L}^{stage2} = \begin{cases}
        \mathcal{L}^{stage1},                 & if  I \in D_{3D} \\
        \mathcal{L}_{2D} + \beta_{fr}\mathcal{L}_{fr} + \beta_{ar}\mathcal{L}_{ar}, & \text{otherwise.}
    \end{cases}
\end{equation}

\section{Experiments}
To evaluate the effectiveness of our proposed framework, we conduct extensive experiments on both Rendered hand pose dataset \cite{LearnToEstimate3DHandPoseFromSingleImage} and Stereo Hand Pose Tracking Benchmark \cite{stereodataset} and then compare with the state-of-the-art methods. We also investigate the performance of our keypoint based hand detector by evaluating the accuracy of hand detection and hand side prediction. Finally, we use weakly-supervised training on a trained 3D hand pose estimation model to show the effects of the different proposed constraints. The following subsections describe the details of the experiments and their results.
%full pipline one hand  and both hand results
\subsection{Datasets}
We use the following datasets for our experiments:

\subsubsection{Rendered hand pose dataset}
Identifying the need of a RGB image dataset for hand pose estimation, \cite{LearnToEstimate3DHandPoseFromSingleImage} developed the Rendered hand pose dataset. This dataset is built using 20 different computer rendered characters performing 39 different actions. The characters are being looked at from varying camera angles and with random backgrounds. The dataset contains 41,258 images for training and 2,728 images for evaluation. For the keypoints, there are in total of 21 keypoints where each finger is represented by 4 keypoints. Again, for each of the keypoints, there is information as to whether they are visible or occluded/cropped in the image. Camera intrinsic parameters, segmentation maps and depth maps are also available.

\subsubsection{Stereo Hand Pose Tracking Benchmark}
Stereo Hand Pose Tracking Benchmark \cite{stereodataset} provides both 2D and 3D annotations for 18,000 stereo pairs. The images are of size $640\times480$. We use 6,000 of the total images for evaluation. The dataset contains only left hand of a person with varying background and lighting conditions.

\subsection{Evaluation Protocols}
For the proposed hand detector and hand pose estimator, we provide the following protocols to evaluate hand detection performance, hand side prediction accuracy and hand keypoints estimation performance.

\subsubsection{Hand Detection}
In this section, we intent to compare the effectiveness of our proposed hand detection method with existing hand detectors such as \textit{HandSegNet} and \textit{SSDHandDetect}. Because the three methods were trained on different types of annotation, namely segmentation maps and keypoints, comparing their predicted bounding box to ground truth bounding boxes from the respective annotation used for their training does not give us clear picture of their performance among them. Instead, the most concrete evidence would come from comparing them in their application in a hand pose estimation pipeline by evaluating whether they can provide any improvements in 2D and 3D hand pose estimation accuracy. Hence, we replace the hand detector part of the full hand pose estimation model in \cite{LearnToEstimate3DHandPoseFromSingleImage} with each of the detectors and report 2D and 3D results using standard keypoints detection metrics such as Mean and Median Endpoint Pixel Error (EPE) and Area under the curve (AUC). The full hand pose estimation model consists of a 2D pose estimation part called \textit{PoseNet} and a separate 3D pose estimation module which allows us to report 2D and 3D results separately.

\subsubsection{Hand Side Accuracy}
Similarly, to show that our hand detection network can, in fact, correctly distinguish between left and right hand, we calculate the percentage of correctly labelled hand sides for largest and smallest hands in the input image from two different datasets. We determine the largest hand in a dataset by comparing the number of pixels in a segmentation mask for each of the two hands in an image. A threshold on output heatmaps of fullbody hands keypoint estimator is used to determine if a hand is detected for a hand side.

\subsubsection{3D Hand Pose Estimation}
We report our 3D hand pose estimation results with and without different constraints using the standard metric of area under the curve (AUC) on the percentage of correct keypoints (PCK) over range of error thresholds. Finally, we also report the effects of different constraints on different joints on each of the $X$, $Y$ and $Z$ components of 3D coordinates.

\begin{figure*}
\centering
\subfloat[]
{
    \includegraphics[width=0.5\linewidth]{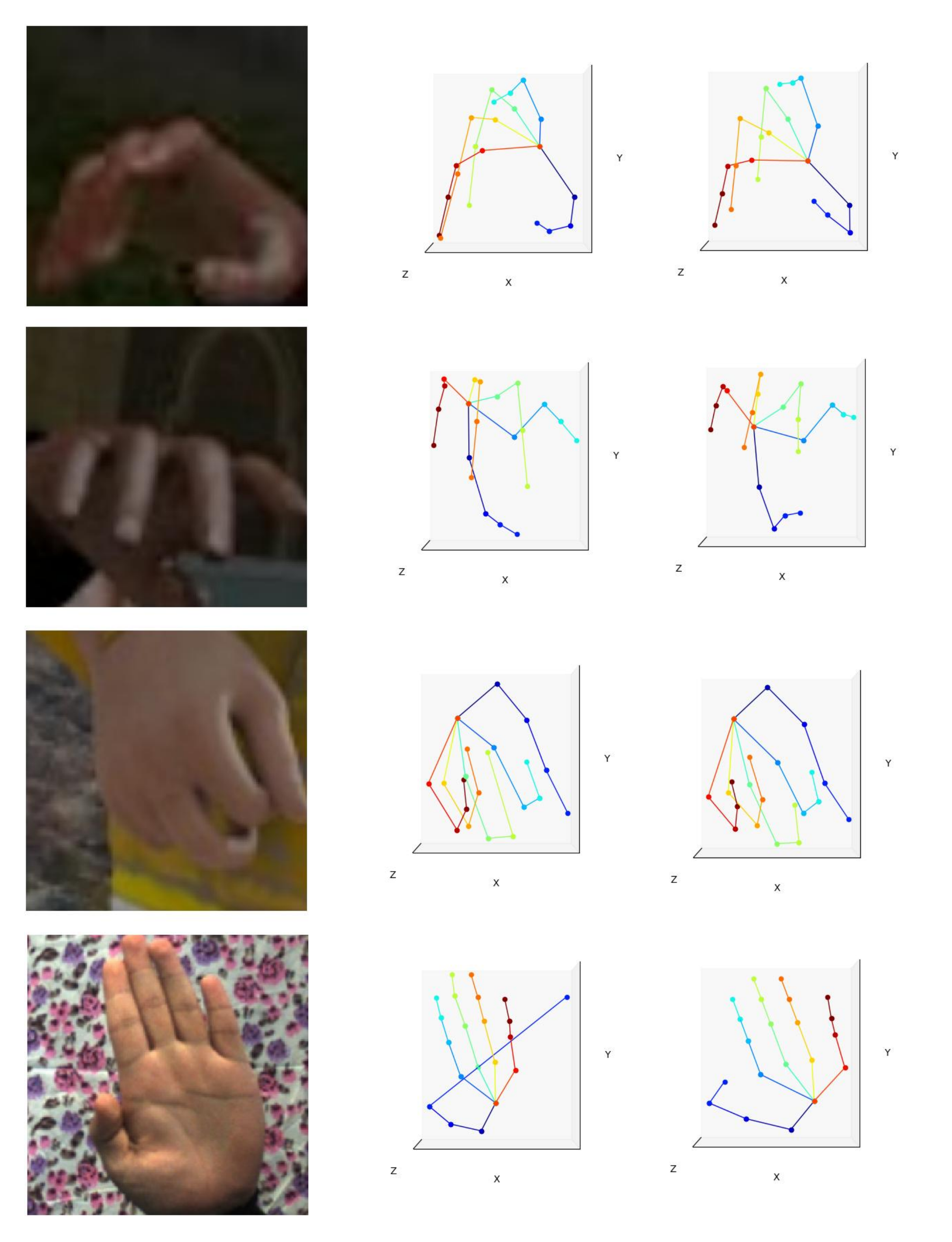}
}
\subfloat[]
{
    \includegraphics[width=0.5\linewidth]{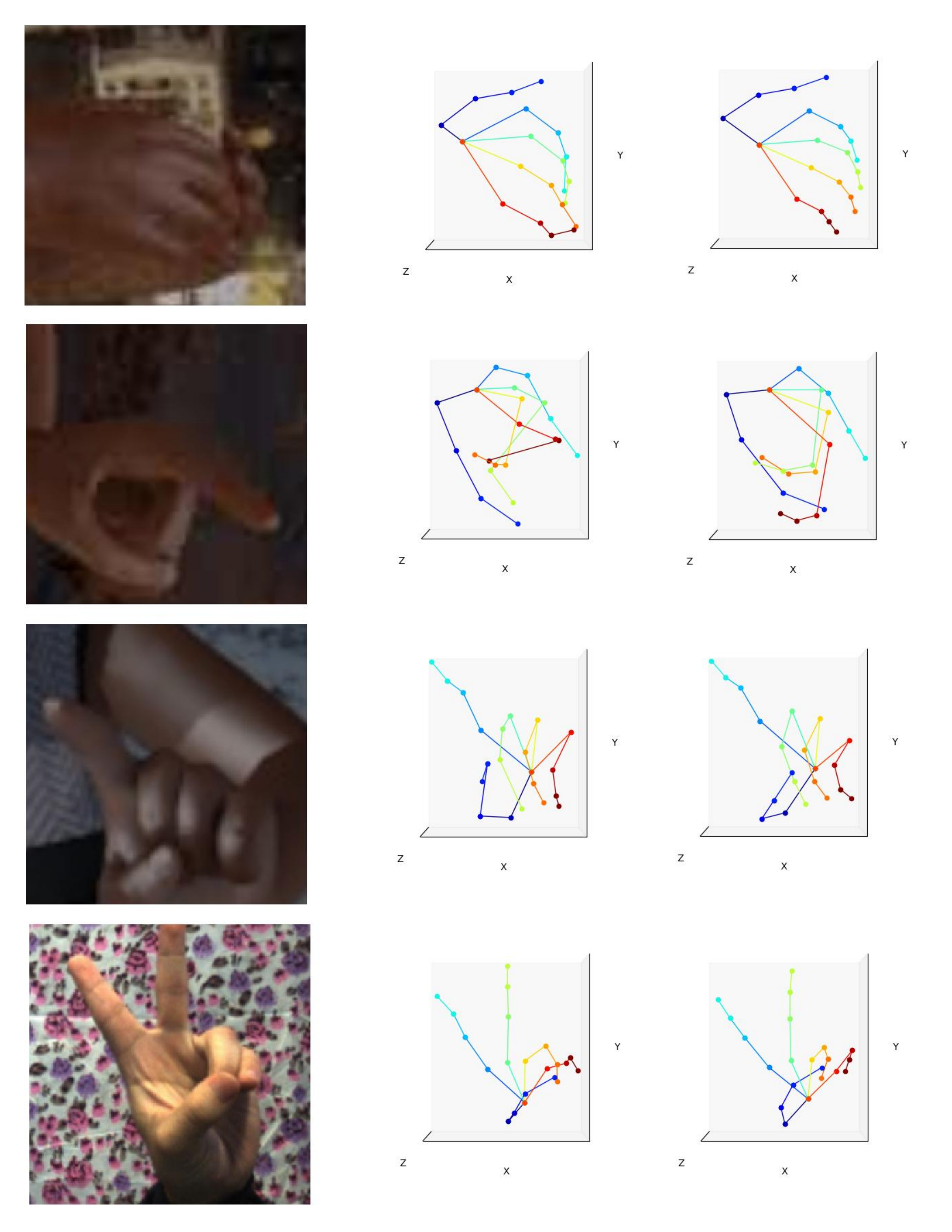}
}
\bigskip
\caption{Images in (a) shows the effect before and then after using relative finger bone length ratio constraint and (b) shows this for angle range constraint.}
\label{fig:CorrectionsMadebyConstraints}
\end{figure*}

\section{Results}
\subsection{Hand Detection}
Table \ref{table:HandSideDetect} shows the accuracy of hand side prediction (with training only on RHD) for the larger and smaller hand (if visible) in different datasets:

\begin{table}[ht]
\centering
\begin{tabular}{c | c c}
\hline\hline %inserts double horizontal lines
 & RHD & STB \\ [0.5ex]
\hline % inserts single horizontal line
Larger hand & 89.8 & 79.8\\
Smaller hand & 73.4 & -\\
\hline %inserts single line
\multicolumn{3}{c}{} \\
\end{tabular}
\caption{Avg. hand side detection accuracy (\%)}
\label{table:HandSideDetect}
\end{table}

As we can see from the table, the accuracy of hand side detection is not only good in the dataset that the network was trained on but also is good in the one it wasn't trained in - i.e. Stereo Hand Pose Tracking Benchmark. This shows good cross-generalization for hand side prediction across different datasets. We cannot provide data for smallest hand in STB because the dataset only consists of images of one hand.

Next, we would like to see whether our hand detector can improve 2D keypoint estimation performance. By using each of the hand detectors with \textit{PoseNet} from \cite{LearnToEstimate3DHandPoseFromSingleImage} trained on manually cropped hand images for 2D hand pose estimation, we get results as shown in Table \ref{table:Hand2D}.

\begin{table}[ht]
\centering
\resizebox{\columnwidth}{!}{\begin{tabular}{c | c c c}
\multicolumn{4}{c}{RHD} \\
\hline\hline %inserts double horizontal lines
 & Mean EPE & Median EPE & AUC\\ [0.25ex]
\hline % inserts single horizontal line
Ours+PoseNet & \textbf{8.922} & \textbf{2.644} & \textbf{0.779}\\
HandSeg+PoseNet & 15.791 & 4.479 & 0.717\\
SSDHandDetect+PoseNet & 37.178 & 10.641 & 0.495\\
\hline
\multicolumn{4}{c}{} \\
\multicolumn{4}{c}{STB} \\
\hline\hline
 & Mean EPE & Median EPE & AUC\\ [0.25ex]
\hline % inserts single horizontal line
Ours+PoseNet & \textbf{6.899} & \textbf{4.946} & \textbf{0.780}\\
HandSeg+PoseNet & 7.321 & 6.320 & 0.765\\
SSDHandDetect+PoseNet & 15.272 & 10.251 & 0.587\\
\hline %inserts single line
\multicolumn{4}{c}{} \\
\end{tabular}}
\caption{2D Hand keypoint estimation in RHD followed by estimation in STB where the models were trained jointly in RHD \& STB training set}
\label{table:Hand2D}
\end{table}

As we can see from the table, our hand detector can consistently provide better results in 2D keypoints estimation. Additionally, we can also see the large difference between Mean and Median EPE values with \textit{HandSegNet+PoseNet} \cite{LearnToEstimate3DHandPoseFromSingleImage} in case of RHD dataset evaluation than our solution. This large discrepancy could be attributed to \textit{HandSegNet} producing a single big crop in cases when two hands are close to each other as shown in Fig. \ref{fig:HandDetect_Samples}.

\begin{figure}
\centering
\includegraphics[scale=0.55]{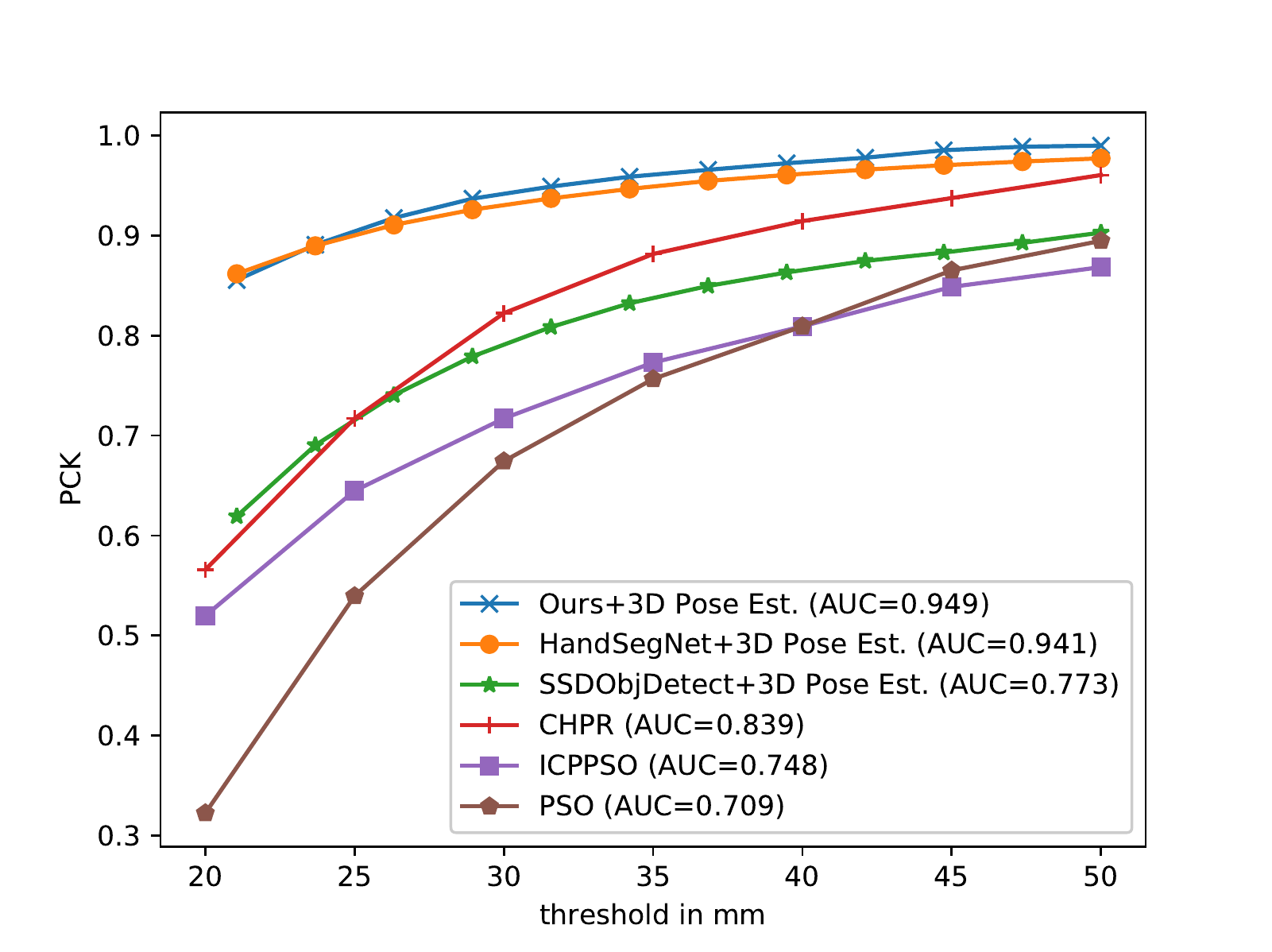}
\caption{3D PCK results on STB dataset's evaluation samples with different hand detectors and PoseNet trained on RHD \& STB jointly.}
\label{fig:HandDetect_AUC3DGraph}
\end{figure}

Finally, we would like to see the effect on 3D pose estimation. So, we setup the hand detectors for a 3D estimation pipeline and evaluate their effect using AUC. Results seen in Fig. \ref{fig:HandDetect_AUC3DGraph} clearly shows that our hand detector has a positive effect on raising the accuracy in 3D hand pose estimation when the results are compared with other hand detection methods.

\begin{figure*}[!tp]
\centering
\minipage{0.5\textwidth}
\includegraphics[width=1.0\linewidth]{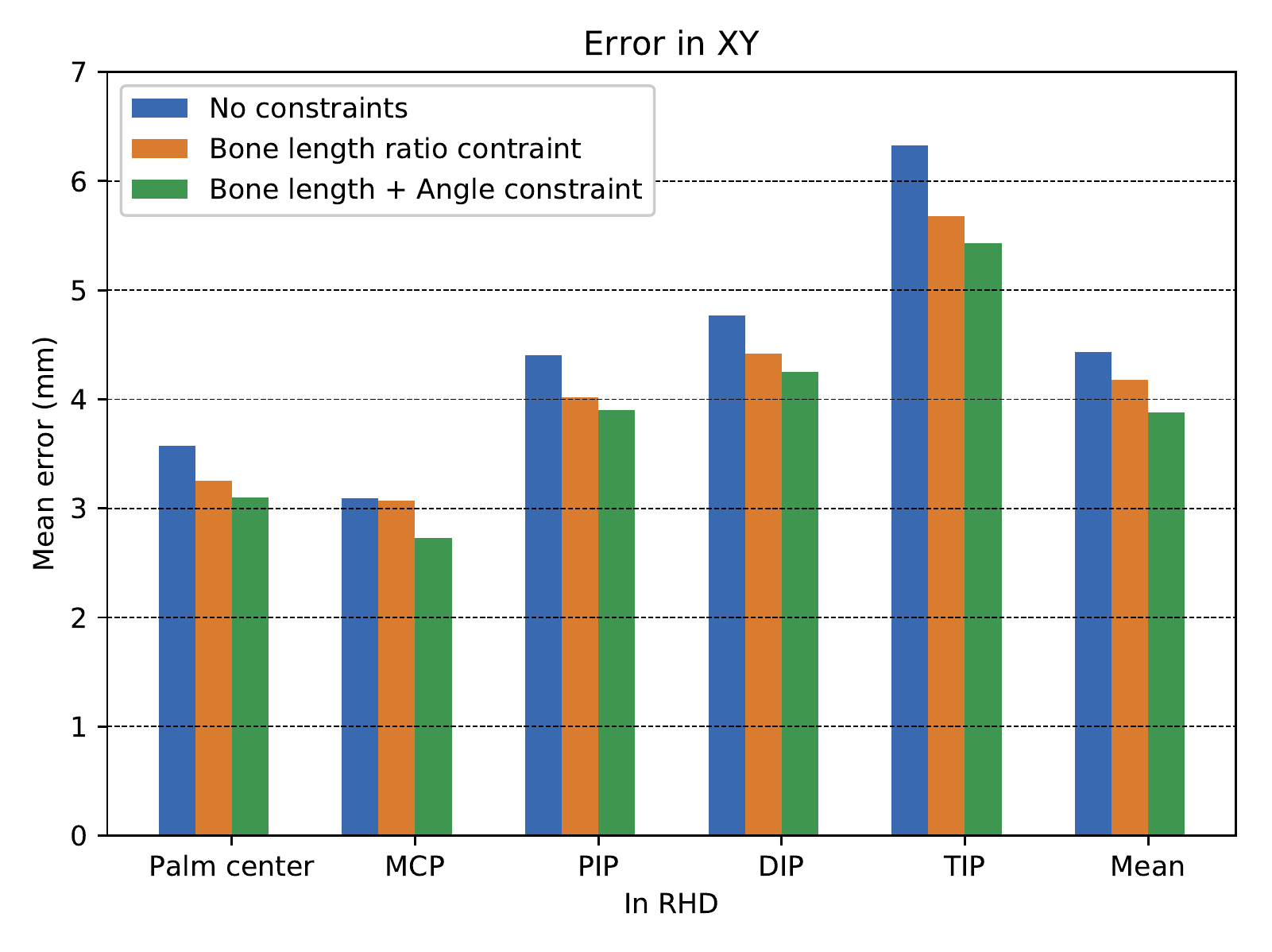}
\endminipage\hfill
\minipage{0.5\textwidth}
\includegraphics[width=1.0\linewidth]{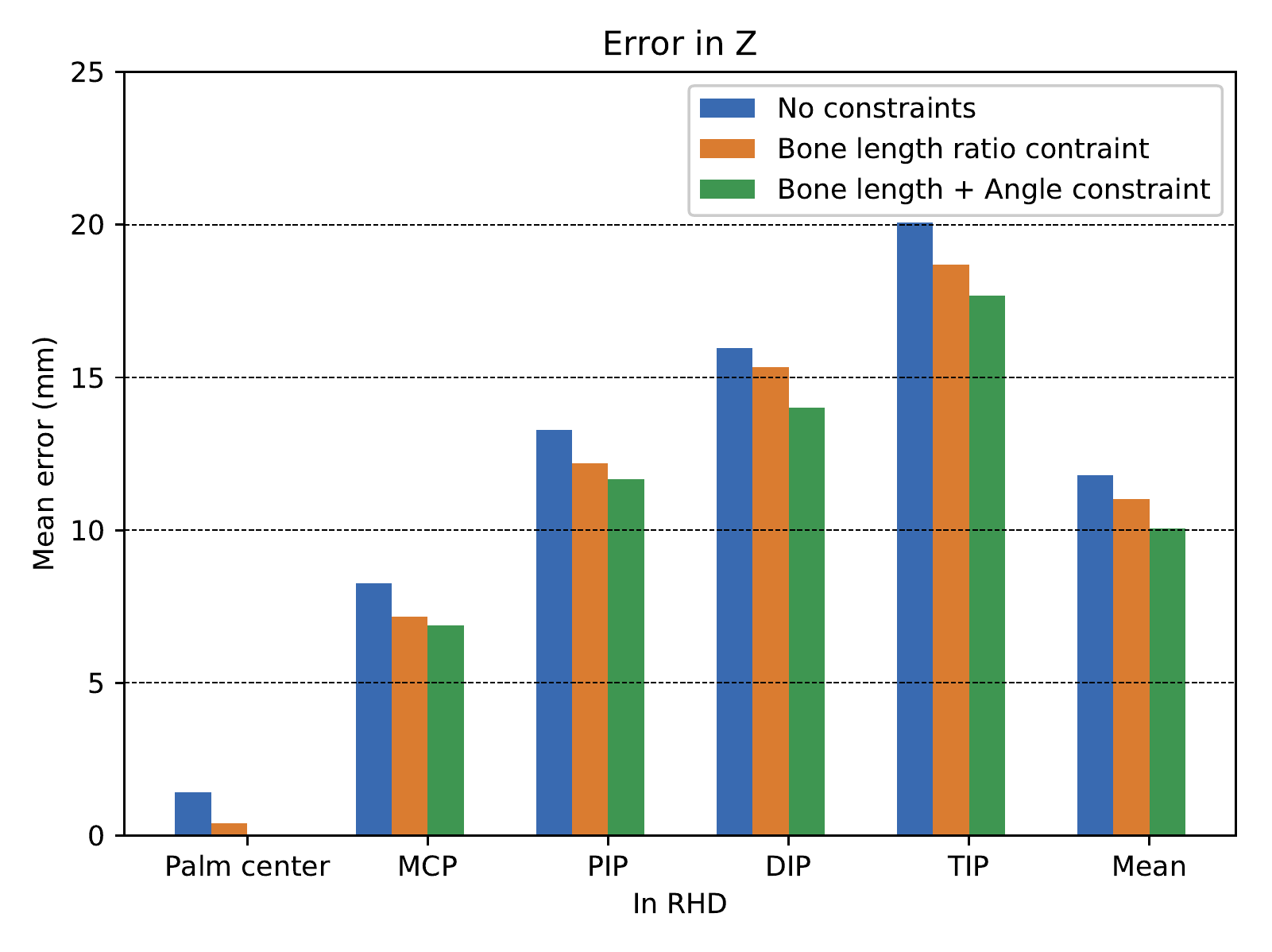}
\endminipage\hfill
\bigskip
\caption{Effect of each of the two proposed constraints on 2D coordinates (X and Y) and depth (Z) components of the predicted 3D coordinates of different joints for \textit{Rendered Hand Pose Dataset}. Errors in different joints decrease by different amounts after applying each constraint.}
\label{fig:MeanErrorsInJoints_XYZ}
%\minipage{0.33\textwidth}
%\includegraphics[width=1.0\linewidth]{MeanErrorX_RHD.pdf}
%\endminipage\hfill
%\minipage{0.33\textwidth}
%\includegraphics[width=1.0\linewidth]{MeanErrorY_RHD.pdf}
%\endminipage\hfill
%\minipage{0.33\textwidth}
%\includegraphics[width=1.0\linewidth]{MeanErrorZ_RHD.pdf}
%\endminipage\hfill
%\bigskip
%\caption{Effect of each of the two proposed constraints on X, Y and Z components of the predicted 3D coordinates of different joints for \textit{Rendered Hand Pose Dataset}. Errors in different joints decrease by different amounts after applying each constraint.}
%\label{fig:MeanErrorsInJoints_XYZ}
\end{figure*}

\subsection{3D Hand Pose Estimation}
We provide 3D hand pose estimation results on different datasets to show the effectiveness of our constraints. In our results, we have shown the performance of our baseline model upon which each of the constraints were applied and how the constraints have helped to improve its performance further. Additionally, we also show results from other methods and how they compare to ours.

But first, we test that the relative sizes of fingers on an average hand across both datasets are in fact similar in relative length to each other. Table \ref{table:HandPose_RelativeFingerLengthsAcrossDatasets} shows that this theory holds within acceptable margin of error given the inherent uncertainty induced with manual 3D annotation of keypoints in STB dataset.
\begin{table}[ht]
\centering
\begin{tabular}{c | c c}
\hline\hline %inserts double horizontal lines
\textbf{Fingers} & RHD & STB \\ [0.5ex]
\hline % inserts single horizontal line
Thumb & 6.4 & 6.0\\
Index & 9.5 & 9.2\\
Middle & 10.0 & 10.0\\
Ring & 9.0 & 8.8\\
Little & 7.4 & 7.0\\
\hline %inserts single 
\multicolumn{3}{c}{} \\
\end{tabular}
\caption{Mean relative finger lengths across datasets when the longest finger (i.e. middle finger) is re-scaled to 10}
\label{table:HandPose_RelativeFingerLengthsAcrossDatasets}
\end{table}

Secondly, we calculate the mean finger bone length ratio for each dataset in their training set using Eq. (\ref{eq:8}). Results are shown in Table \ref{table:HandPose_MeanRelativeFingerLengthRatioAmongDatasets}.
\begin{table}[ht]
\centering
\begin{tabular}{c | c c}
\hline\hline %inserts double horizontal lines
 & RHD & STB \\ [0.5ex]
\hline % inserts single horizontal line
Mean & 1.08839 & 1.13241\\
Variance & 0.00144 & 0.00055\\
\hline %inserts single line
\multicolumn{3}{c}{} \\
\end{tabular}
\caption{Mean relative finger bone length ratios and their variances in different datasets}
\label{table:HandPose_MeanRelativeFingerLengthRatioAmongDatasets}
\end{table}

Similarly, we calculate angle ranges, consisting of minimum and maximum angles, from training set of each dataset as shown in Table \ref{table:HandPose_DigitAngleRangesAmongDatasets}. 
\begin{table}[ht]
\centering
\begin{tabular}{c | c c}
\multicolumn{3}{c}{RHD} \\
\hline\hline %inserts double horizontal lines
 & Min & Max \\ [0.5ex]
\hline % inserts single horizontal line
Middle and Proximal phalanx & 0.33 & 91.95\\
Distal and Middle phalanx & 0.21 & 125.39\\
Index-Middle-Ring Flexion & 20.88 & 108.63\\%3.36 & 139.73\\
\hline %inserts single line
\end{tabular}

\begin{tabular}{c | c c}
\hline
\multicolumn{3}{c}{} \\
\multicolumn{3}{c}{STB} \\
\hline\hline %inserts double horizontal lines
 & Min & Max \\ [0.5ex]
\hline % inserts single horizontal line
Middle and Proximal phalanx & 0.08 & 90.00\\
Distal and Middle phalanx & 0.08 & 90.00\\
Index-Middle-Ring Flexion & 21.90 & 102.03\\
\hline %inserts single line
\multicolumn{3}{c}{} \\
\end{tabular}
\caption{Mean angle ranges of different digits in different datasets (DEG)}
\label{table:HandPose_DigitAngleRangesAmongDatasets}
\end{table}

\begin{figure}
\centering
\includegraphics[scale=0.50]{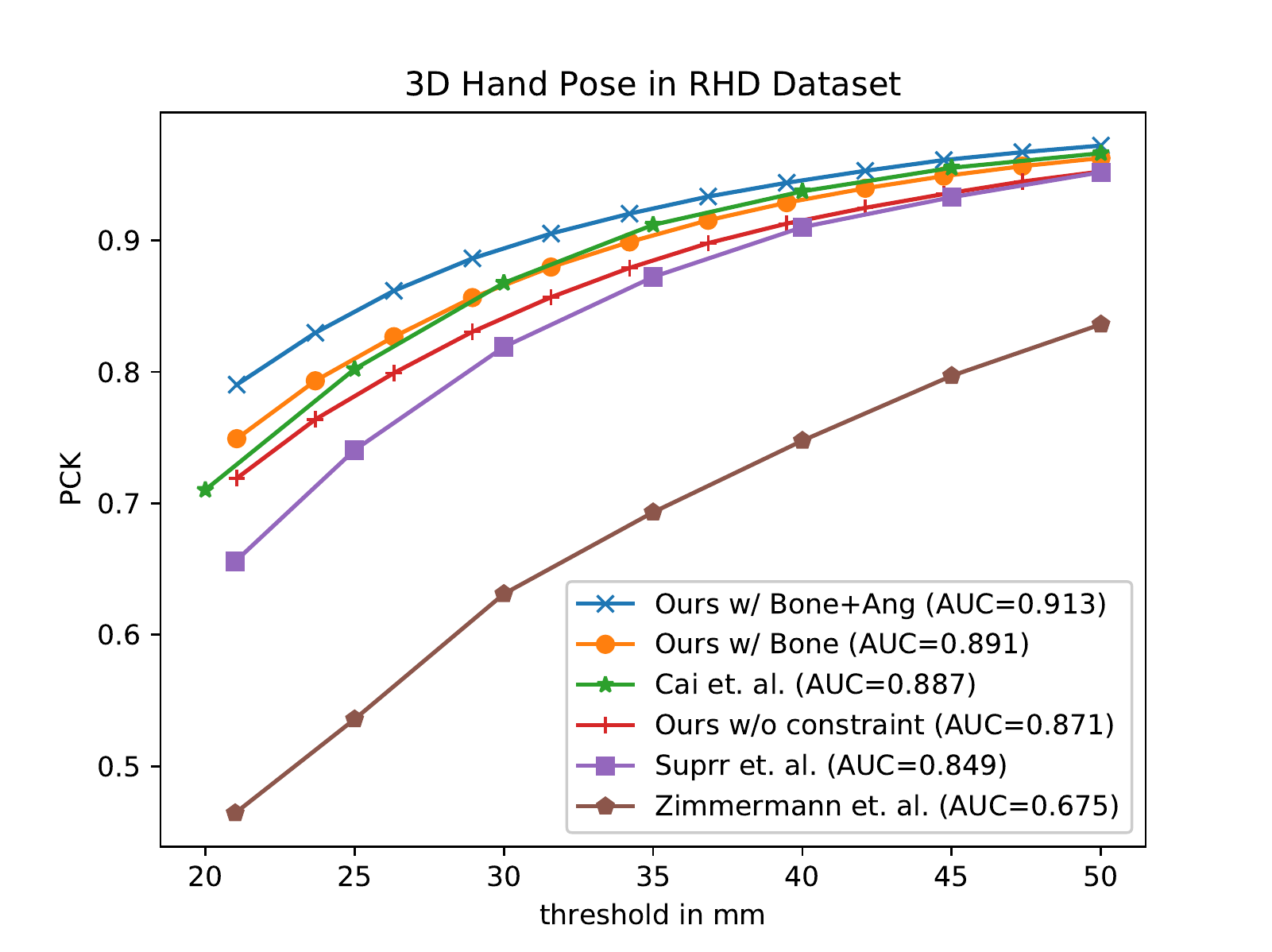}
\caption{3D PCK results on RHD dataset's evaluation samples using our constraints.}
\label{fig:HandPose_3DRHD}
\end{figure}

Using the values in Table \ref{table:HandPose_MeanRelativeFingerLengthRatioAmongDatasets} and \ref{table:HandPose_DigitAngleRangesAmongDatasets} we train our 3D Hand pose estimation model in a weakly-supervised setting. Fig. \ref{fig:HandPose_3DRHD} shows how our proposed constraints are effective in increasing performance for \textit{Rendered Hand Pose} dataset. Even though \cite{Weakly3DHandPose} uses extra annotation of depth maps for training, we are able to outperform by a good margin after making use of all our proposed constraints.

\begin{figure}
\centering
\includegraphics[scale=0.50]{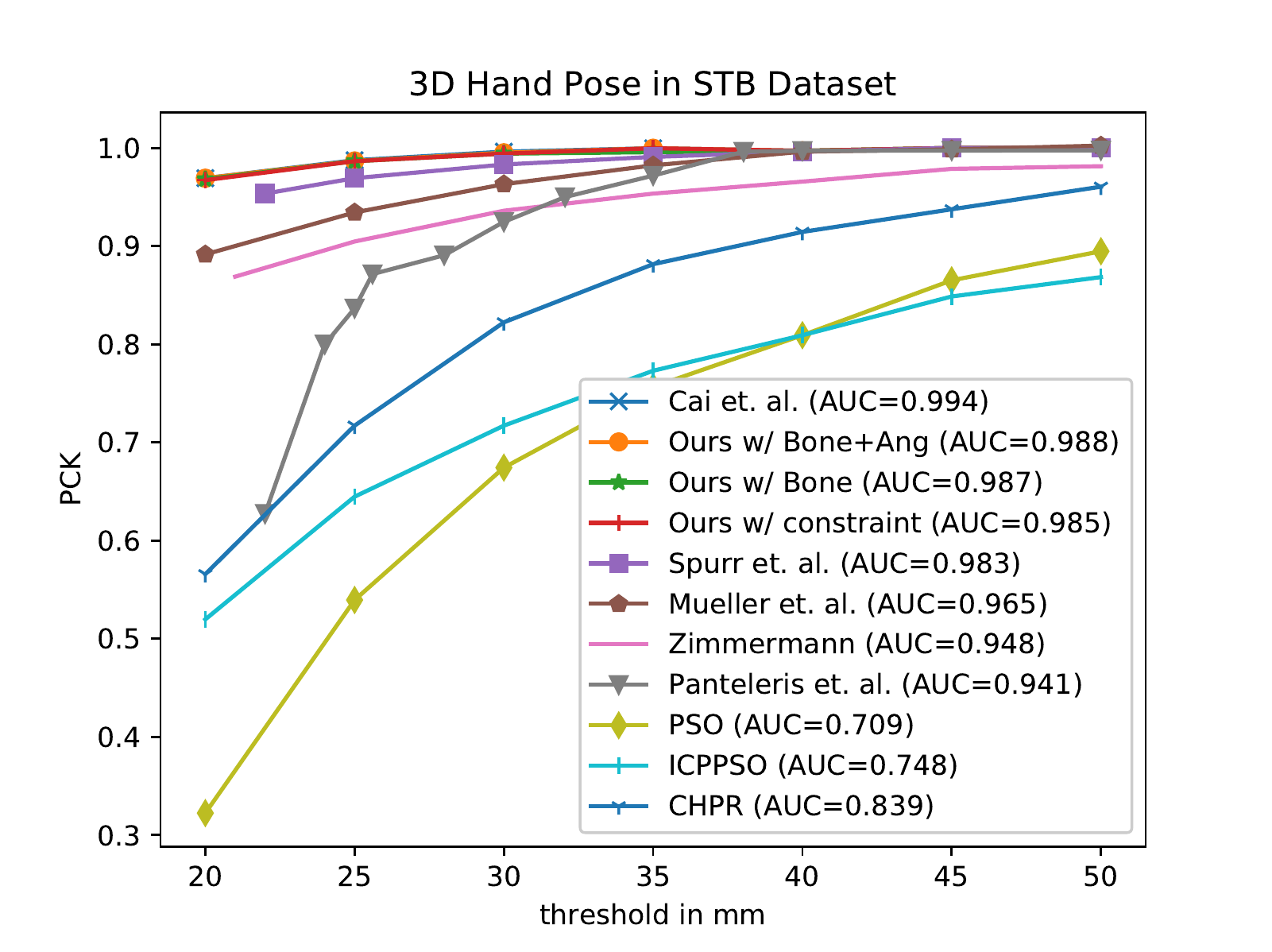}
\caption{3D PCK results on STB dataset's evaluation samples using our constraints.}
\label{fig:HandPose3D_STB}
\end{figure}

Similarly, Fig. \ref{fig:HandPose3D_STB} shows how our constraint can further improve the performance on our baseline model for STB dataset. In this dataset, though, we are not able to outperform \cite{Weakly3DHandPose} even though our method gives better results than many other hand pose estimation methods. We attribute this to the fact that the extra depth map annotation provided their model with better information to aid in learning 3D hand structure. Furthermore, Table \ref{table:HandPose_STBMeanEPE} shows the comparison of Mean Average Endpoint Pixel Error with \cite{3DHandShapePoseInWild} in STB dataset. Again, we see that our method produces a slightly better results in 3D prediction.

\begin{table}[ht]
\centering
\begin{tabular}{c | c c}
\hline\hline %inserts double horizontal lines
 & Ours & \cite{3DHandShapePoseInWild} \\ [0.5ex]
\hline % inserts single horizontal line
Mean EPE & 8.71 & 9.76\\
\hline %inserts single line
\end{tabular}
\caption{Mean EPE for 3D prediction to ground truth in STB (mm)}
\label{table:HandPose_STBMeanEPE}
\end{table}

\hfill
\break
\noindent
\textbf{Ablation Studies on Constraints}
To see the full effect after the application of each constraint, we calculate and plot the average joint prediction errors in the 2D (i.e. X and Y) and depth component (i.e. Z) of different position/joint 3D keypoint coordinates from RHD dataset. We use the following position/joints for our study - Palm center, Metacarpophalangeal (MCP), Proximal interphalangeal (PIP), Distal interphalangeal (DIP) and finger tip (TIP). Fig. \ref{fig:MeanErrorsInJoints_XYZ} shows that the proposed constraints do indeed have a positive impact on improving predictions by a good margin. The constraints not only aid in depth prediction, as expected, but also seem to help with 2D prediction. Again, we can also see that the effects of the constraints are more prominent as we take the positions/joints further from palm center. Joints/positions around easily located position like palm center are themselves relatively more easy to predict for deep learning models. But as we take joints/position further from palm center, they become iteratively harder to locate. Our proposed constraints seem to aid the models especially in these scenarios by providing hand geometry-based supervision. Finally, the improvements seen from finger bone length constraint seem to almost always out weigh the improvements from bone length and angle constraint together. This may suggest that the bone length constraint also corrects some invalid angles along with invalid bone lengths and they overlap in aiding to correct some types of invalid hand poses.

For a more visual study of these effects, we pick select samples from the datasets which prominently show the correcting effect of the constraints. The samples are shown in Fig. \ref{fig:CorrectionsMadebyConstraints}. Images in Fig. \ref{fig:CorrectionsMadebyConstraints}(a) shows how relative finger bone length ratio constraint can correct abnormally long digit length predictions to a more plausible digit position. Similarly, Fig. \ref{fig:CorrectionsMadebyConstraints}(b) shows how angle range constraint can correct impossible digit bends to a more acceptable angle for that digit pair.

\section{Conclusion}
In this paper, we presented a complete pipeline for hand pose estimation that accomplishes all tasks of hand detection, hand side prediction and hand pose estimation required for a real-world implementation of a complete hand pose pipeline. Firstly, we addressed the common problem of ambiguity faced by hand detectors in confusing backgrounds and hand adjacent condition by using a keypoints-based approach to hand detector. We showed experimentally how this method can avoid such problems and improve performance of any ad-hoc hand estimation network attached to it. Similarly, we also discussed how we can make use of biological hand constraints of finger bone length ratio and angle ranges to create losses which can aid in further improving a hand estimation network. Finally, we provided experimental results showing improvements on our baseline methods after using these constraints and provided comparison of our method against other state-of-art 3D hand estimation methods.

\section{Future Work}
In the future, we would like to extend the work done our design for hand detection by including supervision with segmentation mask. This increased supervision should improve the performance of hand detector even more. Furthermore, we would like to redesign the architecture such that both hand detector and hand pose estimator can be trained in a joint manner where the hand detector can automatically zoom in to a single hand in an incremental manner.

\ifCLASSOPTIONcaptionsoff
  \newpage
\fi

\bibliographystyle{IEEEtran}
\bibliography{handpose}

\end{document}